\documentclass[10pt,twocolumn,letterpaper]{article}

\usepackage[dvipsnames,table]{xcolor}
\usepackage[]{iccv}      

\definecolor{iccvblue}{rgb}{0.21,0.49,0.74}
\usepackage[pagebackref,colorlinks,allcolors=iccvblue]{hyperref}

\usepackage{pifont}
\usepackage{listings}
\usepackage{marvosym}
\usepackage{textcomp}
\usepackage{amsmath, amssymb} 

\def\paperID{958} 
\def\confName{ICCV}
\def\confYear{2025}

\title{OwlSight: A Robust Illumination Adaptation Framework for Dark Video Human Action Recognition}

\author{%
Shihao Cheng$^{*1}$ \quad Jinlu Zhang$^{*2}$ \quad Yue Liu$^{1}$ \quad Zhigang Tu$^{1\,\dagger}$\\
$^1$Wuhan University \quad
$^2$Peking University \\
\vspace{-20pt}
}

\date{} 
\begin{document}
\twocolumn[{
\renewcommand\twocolumn[1][]{#1}
\maketitle
\begin{center}
    \captionsetup{type=figure}
    \includegraphics[width=\textwidth]{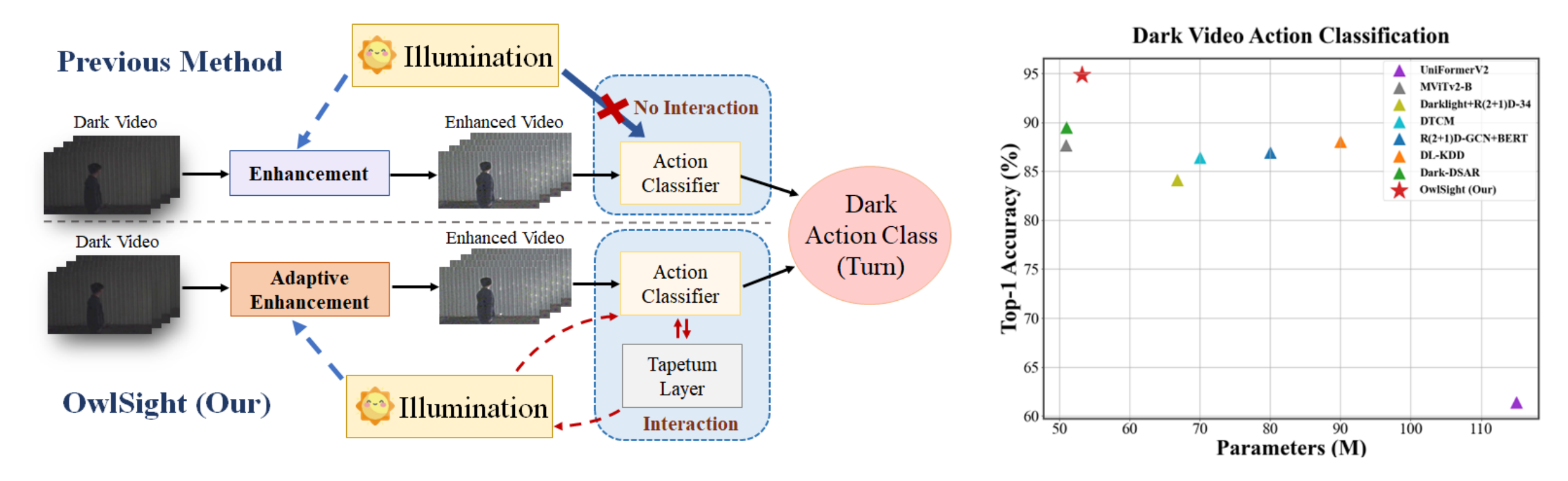}
    \captionof{figure}{\textbf{Left:} Overview of our OwlSight. In contrast to the previous methods, we place a strong emphasis on fully exploring and utilizing illumination information throughout all stages interactively to boost the performance of dark video action recognition. \textbf{Right:} Accuracy (Top-1 Accuracy) and efficiency (parameters) comparison with the state-of-the-arts on the ARID1.5~\cite{xu2021arid} dataset.}
    \label{fig:over}
\end{center}
}]

\begin{abstract}
Human action recognition in low-light environments is crucial for various real-world applications. However, the existing approaches overlook the full utilization of brightness information throughout the training phase, leading to suboptimal performance. To address this limitation, we propose OwlSight, a biomimetic-inspired framework with whole-stage illumination enhancement to interact with action classification for accurate dark video human action recognition. Specifically, OwlSight incorporates a Time-Consistency Module (TCM) to capture shallow spatiotemporal features meanwhile maintaining temporal coherence, which are then processed by a Luminance Adaptation Module (LAM) to dynamically adjust the brightness based on the input luminance distribution. Furthermore, a Reflect Augmentation Module (RAM) is presented to maximize illumination utilization and simultaneously enhance action recognition via two interactive paths. Additionally, we build \textbf{Dark-101}, a large-scale dataset comprising 18,310 dark videos across 101 action categories, significantly surpassing existing datasets (e.g., ARID1.5 and Dark-48) in scale and diversity. Extensive experiments demonstrate that the proposed OwlSight achieves the state-of-the-art performance across four low-light action recognition benchmarks. Notably, it outperforms prior best approaches by \textbf{5.36\% on ARID1.5} and \textbf{1.72\% on our Dark-101}, highlighting its effectiveness in challenging dark environments. 

\renewcommand{\thefootnote}{\fnsymbol{footnote}} 
\footnotetext[1]{Shihao Cheng and Jinlu Zhang contributed equally to this work.}
\footnotetext[2]{Zhigang Tu is the corresponding author (\href{mailto:tuzhigang@whu.edu.cn}{tuzhigang@whu.edu.cn}).}
\renewcommand{\thefootnote}{\arabic{footnote}} 

\end{abstract}

\section{Introduction}
\label{sec:intro}

Recognizing human actions in dark videos is a useful yet challenging visual task in real-world scenarios. It aims to identify human actions in conditions of low brightness and low contrast, which can benefit a large range of applications, such as video security surveillance, vision-based autonomous driving systems, etc. 

Given its practical significance, numerous methods have been proposed for action recognition in low-light videos. However, these methods have certain limitations. Some studies seek to enhance action recognition in low-quality videos by integrating sensor data beyond RGB video. For example, \cite{zhu2020eemefn} uses infrared data features to improve action recognition accuracy in dark environments. However, the incorporation of additional sensor data significantly increases costs, complicates deployment configurations, and adds to processing complexity. On the other hand, augmentation-based methods~\cite{9879165} are also widely used, where they typically apply Low-Light Video Enhancement (LLVE) to boost the visual quality of low-illumination videos before utilizing the enhanced video frames for action recognition. However, LLVE methods often depend on optical flow, coming with high computational cost. Alarmingly, the prior research \cite{xu2021arid} has shown that most illumination enhancement techniques do not improve action recognition performance in dark videos, as their task coherence tends to be fragmented.

Recently, one-stage low-light video action recognition methods have been presented \cite{tu2023dtcm}, allowing the entire network to be trained in a single phase. 
However, these methods primarily focus on illumination enhancement as a preprocessing step, rather than integrating illumination adaptation directly into the training process, which limits their effectiveness in enhancing recognition performance.

\begin{figure}[ht] 
    \centering
    \includegraphics[width=0.45\textwidth]{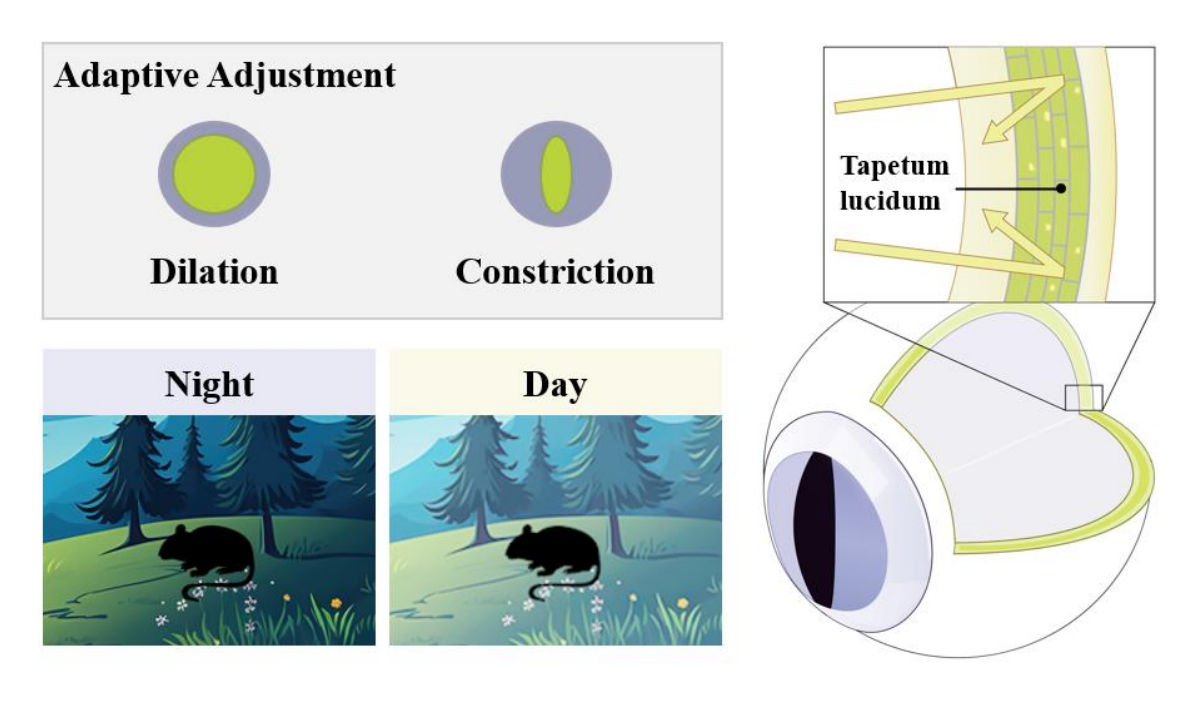}
    \vspace{-0.4cm}
    \caption{ \textbf{Left:} Nocturnal animals dilate pupils at night for capturing enhanced light and constrict them during the day to avoid overexposure.
        \textbf{Right:} The tapetum lucidum, a reflective retinal layer, reflects light to improve illumination and visual clarity.}
    \label{fig:1}
\end{figure}

Studying night vision in biological systems~\cite{science2024} reveals that creatures with enhanced night vision possess unique visual adaptations. As illustrated in Figure~\ref{fig:1}, in dim light, the pupil adjusts its size based on brightness, regulating light entry. Unabsorbed light is then reflected by the tapetum, a layer behind the retina, providing a second opportunity for light capture and improving visual clarity.

Drawing inspiration from this biological mechanism, we propose OwlSight, an end-to-end trained model designed for action recognition in low-light environments, as shown in Figure~\ref{fig:over}. Firstly, the lightweight Time-Consistency Module (TCM) is designed to capture shallow spatiotemporal features. Secondly, these features are processed by an exploited Luminance Adaptation Module (LAM), which adaptively adjusts global brightness based on the input illumination distribution. Thirdly, the adjusted features are passed to a Reflect Augmentation Module (RAM), which processes the input features to generate a `reflected’ version, subsequently handled via a specialized encoder. By integrating the reflected features with the input features across multiple stages, RAM maximizes the usage of the available illumination information. 

Noticeably, as models scale up, existing datasets like ARID~\cite{xu2021arid} and ARID1.5~\cite{xu2021arid} are constrained in small-scale. Dark-48~\cite{tu2023dtcm} contains many redundant videos and lacks action-specific content. Synthetic datasets like UAVHuman-Night~\cite{li2021uav} do not capture the natural vision and uneven lighting that are presented in real low-light environments. To address these issues, we constructed a new large-scale dataset called Dark-101, which is composed of 15,621 low-light videos spanning 101 action categories in 138 diverse scenes.

Our key contributions are summarized as follows:
\begin{itemize}
     \item We propose \textbf{OwlSight}, a biomimetic-inspired framework to fully exploit illumination information for dark action recognition in the whole training stage interactively.
    
    \item We introduce a \textbf{Luminance Adaptation Module (LAM)} and a \textbf{Reflect Augmentation Module (RAM)}, where LAM adaptively adjusts the global brightness based on the input illumination distribution, RAM generates a reflected version of the input features and merges them with the original features to leverage illumination interactively.
    
    \item We present \textbf{Dark-101}, a large-scale low-light video dataset composed of 18,310 samples across 101 action categories, more than 2 times bigger than the previous ones (18310 vs. 8815), enabling to train larger dark video action recognition models.

    \item Extensive experiments demonstrate that our OwlSight achieves the state-of-the-art performance across four low-light action recognition benchmark datasets, e.g. it surpasses the previous methods by \textbf{5.36\%} on the widely-used dataset ARID1.5~\cite{xu2021arid}.
\end{itemize}

\section{Related Work}
\label{sec:formatting}

\begin{figure*}[ht]
    \centering
    \includegraphics[width=1\textwidth]{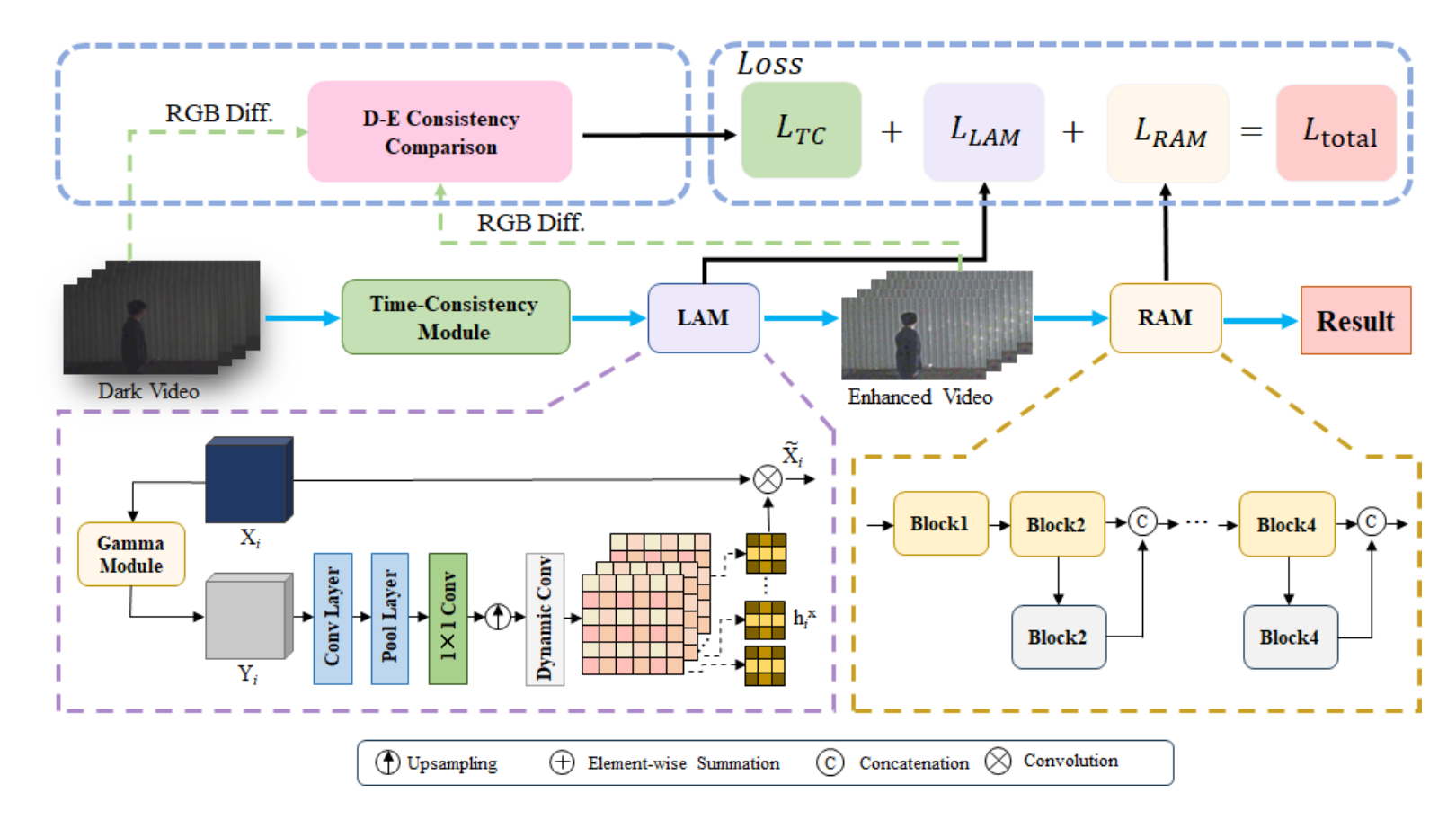} 
    \caption{\textbf{The Overall Architecture of OwlSight.} OwlSight consists of three key components: the Time-Consistency Module (TCM), the Luminance Adaptation Module (LAM), and the Reflect Augmentation Module (RAM). The TCM extracts shallow spatiotemporal features to preserve temporal coherence across the sequence. The LAM adaptively adjusts global brightness based on the input luminance distribution. The RAM further enhances available illumination meanwhile improving action recognition through two interactive pathways.
    The Dark-Enhanced (D-E) consistency of the video frames, which is beneficial for video action recognition in low-light environments, can be preserved by comparing the RGB-difference before and after dark video enhancement.
}
    \label{fig:2}
\end{figure*}

\subsection{Low-light Image and Video Enhancement}
Existing image enhancement methods can be divided into traditional methods and deep learning-based methods. Traditional methods include histogram equalization, gamma correction, and dark light enhancement techniques based on the Retinex theory. Deep learning-based methods can be classified into several categories: 1) Zero-shot learning-based methods, which do not rely on paired training data, such as Zero-DCE~\cite{1guo2020zero} and RRDNet~\cite{zhu2020zero}. Zero-DCE~\cite{1guo2020zero} transforms the low-light enhancement problem into an image-specific curve estimation problem, achieving image enhancement via a non-reference loss function. 2) Reinforcement learning-based methods, which utilize deep reinforcement learning for exposure optimization. For example, DeepExposure~\cite{yu2018deepexposure} automatically learns local exposure for each sub-image meanwhile globally designs rewards to balance overall exposure. 3) Unsupervised learning-based methods, which also do not require paired training data. E.g., EnlightenGAN~\cite{jiang2021enlightengan} enhances low-light images using GANs, and PairLIE~\cite{fu2023learning} uses paired weak-light instances for training. 4) Supervised learning-based methods, which require a large amount of labeled data for training, like~\cite{yan2024you, cai2023retinexformer, wang2019underexposed, zhu2020eemefn, lim2020dslr, lu2020tbefn, wang2020lightening, triantafyllidou2020low}. CIDNet~\cite{yan2024you} introduces HVI color space and enhances images through a dual-path approach. Retinexformer~\cite{cai2023retinexformer} combines Retinex theory with transformers. However, directly applying low-light enhancement to videos can lead to temporal consistency issues, which is why LLVE was proposed. MBLLVEN~\cite{lv2018mbllen} extracts temporal information using 3D convolutions to maintain temporal consistency.  StableLLVE~\cite{li2023fastllve} maintains temporal consistency in enhanced videos by learning and inferring motion fields (i.e. optical flow) from synthesized short-distance video sequences. Our method also considers the correlation and consistency across temporal dimension, where we present a lightweight slow-fast time consistency module to effectively ensure temporal smoothness in video.

\subsection{Dark video Action Recognition}
Initially, conventional action recognition methods, such as C3D~\cite{tran2015learning}, P3D ResNet~\cite{qiu2017learning}, and I3D~\cite{carreira2017quo}, were introduced to tackle visual tasks under low-light conditions. However, these methods are not well-suited for challenging dark environments. Existing approaches for action recognition in dark videos can be broadly categorized into three types: modality-assisted methods, two-stage augmentation methods, and one-stage augmentation methods.

Modality-assisted methods boost action recognition in low-quality video with the help of some sensor information other than RGB video. Ulhaq~\cite{ulhaq2018action} fused multiple video stream deep features corresponding to multiple spectra to combine visible light sequences with infrared data features in night vision scenes. For augmentation-based two-stage methods, DarkLight~\cite{chen2021darklight} enhances videos using gamma correction and applies action recognition through a dual-path approach. The two-stream action recognition method~\cite{suman2023two} builds on this dual-path framework, using self-calibrated supervision (SCI)~\cite{tran2018closer}with temporal graph encoding for action recognition. For augment-based one-stage methods, DTCM~\cite{tu2023dtcm} achieves end-to-end learning and introduces a dedicated dark video dataset, Dark48. DSAR~\cite{yin2024dark} extends this structure by integrating MViTv2~\cite{li2022mvitv2} with dark enhancement techniques, yielding further improvements.

To achieve more efficient and accurate action recognition, our model, OwlSight, integrates a nocturnal vision mechanism with a Transformer-based architecture, enabling flexible capture of multi-scale spatiotemporal features in low-light scenes, and interactively utilizing illumination information through dual pathways.

\subsection{Low-Light Action Recognition Datasets}
Low-light video action recognition has been advanced by several key datasets. ARID~\cite{xu2021arid} is one of the earliest datasets, containing 3,784 video clips across 11 action categories, all captured in dark environments. However, its limited size restricts model training. To address this, ARID1.5~\cite{xu2021arid} expands the dataset by adding more action categories and varying lighting conditions, resulting in a total of 5,572 video clips. Nonetheless, the dataset remains relatively small, and the individual video durations are brief. Subsequently, Dark-48~\cite{tu2023dtcm} offers a larger and more diverse collection, with 48 action categories and 8,815 video clips, enhancing testing conditions for low-light scenarios. However, the limited number of categories still poses challenges for comprehensive model training. UAVHuman-Night~\cite{li2021uav} focuses on aerial videos captured by drones, including both daytime and nighttime sequences, which is particularly beneficial for low-light surveillance applications.

To address the limitations related to dataset size, category diversity, short video duration, etc., we design Dark-101, a high-quality dataset that includes 15,621 video clips across multiple scenes with longer video durations, enabling support more generalized and robust low-light action recognition investigation.

\section{Proposed Method}

As shown in Fig.~\ref{fig:2}, OwlSight is composed of three main components: the Temporal-Consistency Module, the Luminance Adaptation Module (LAM), and the Reflective Augmentation Module (RAM). Our method first employs the Temporal-Consistency Module to extract shallow spatiotemporal information. Then, LAM adaptively utilizes available illumination information, and RAM further enhances illumination through two interactive pathways, progressively benefiting recognition performance. The entire network is trained end-to-end, effectively integrating both enhancement and recognition.

\subsection{Time-Consistency Module}
\begin{figure}[ht]
    \centering
    \includegraphics[width=0.5\textwidth]{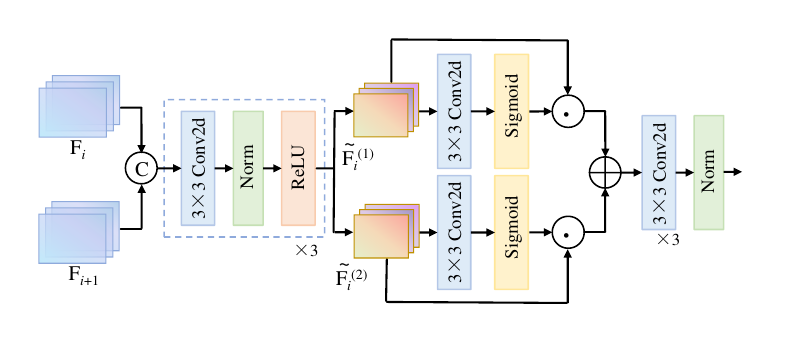} 
    \caption{Architecture of the Time-Consistency Module.}
    \label{fig:3}
\end{figure}
 
The architecture of the Time-Consistency Module is illustrated in Fig.~\ref{fig:3}. For a low-light input video $\{F_i\}_{i=1}^T \in \mathbb{R}^{T \times 3 \times H \times W}$, where $T$ denotes the number of video frames, we first concatenate adjacent frames $F_i$ and $F_{i+1}$ to obtain the paired sequence $\{(F_i, F_{i+1})\}$. This sequence is initially processed by a shallow convolutional network with a kernel size of $3 \times 3$ and a stride of 1, followed by a normalization layer and a ReLU activation function. This processing sequence is repeated three times to progressively extract shallow features $\tilde{F}_i$, which has a spatial dimension of $H \times W$ and a channel dimension of $2C$. Next, we split $\tilde{F}_i$ into two separate components, $\tilde{F}_i^{(1)}$ and $\tilde{F}_i^{(2)}$, along the channel dimension on average. The Inter-frame features $z$ are obtained as:
\begin{equation}
z = \tilde{F}_i^{(1)} \cdot \sigma(W_{3 \times 3}(\tilde{F}_i^{(1)})) + \tilde{F}_i^{(2)} \cdot \sigma(W_{3 \times 3}(\tilde{F}_i^{(2)})),
\end{equation}
where $W_{3 \times 3}(\cdot)$ denotes the $3 \times 3$ convolution operation, and $\sigma$ represents the sigmoid activation function. The outputs $z$ are then passed through three $3 \times 3$ convolutional layers followed by normalization for subsequent feature mixing. The spatiotemporal features $X_i$ are subsequently extracted and fed into the LAM for enhancement.

To preserve temporal coherence across the frame sequence, we adopt the meticulously crafted loss functions from DTCM \cite{tu2023dtcm}, which are defined as follows:
\begin{equation}
L_{\text{TC}} = \frac{1}{T - 1} \sum_{t = 1}^{T - 1} L_{\text{SCF}} \left( \left| D Y_{t + 1} - D Y_t \right|, \left| D I_{t + 1} - D I_t \right| \right),
\end{equation}
where $T$ is the number of input video clip duration, $L_{\text{SCF}}$ measures the discrepancy between the temporal variations of the enhanced and input video frames. $|DY_{t+1} - DY_t|$ and $|DI_{t+1} - DI_t|$ represent the RGB-Difference of the enhanced frame and the dark frame, respectively. See the Supplemental Materials for more details.

\subsection{Luminance Adaptation Module}
We propose the Luminance Adaptation Module (LAM), inspired by the visual adaptation mechanisms observed in nocturnal animals. Many nocturnal species possess the ability to adaptively adjust the response of photoreceptor cells in their retinas, thereby enhancing visual acuity in low-light environments. Similarly, the LAM is designed to dynamically adjust pixel-wise filters in response to local luminance variations in the input image, improving feature extraction under varying lighting conditions.

Specifically, the input feature map \( X_i \) first undergoes a Gamma Module to adjust its luminance mean. This adjustment is achieved by calculating the luminance mean \( \mu_{\text{in}} \) of the input image and applying a Gamma transformation to scale the image's luminance. The Gamma value \( \gamma \) is computed using the following formula:

\begin{equation}
\gamma = \frac{\log(\mu_{\text{out}})}{\log(\mu_{\text{in}})},
\end{equation}
where $\mu_{\text{out}}$ is the target luminance mean and $\mu_{\text{in}}$ is the luminance mean of the input image. Based on the experiments, we set $\mu_{\text{out}} = 0.5$. The details of the experiments can be found in Section 4.3. We initially normalize the input features $X_i$ to the range $[0,1]$ to obtain $X'_i$. Subsequently, the formula for conducting the illumination transformation on the normalized input features $X'_i$ is presented as follows:
\begin{equation}
Y_i = X_i'^{\gamma}.
\end{equation}
The adjusted image is passed through a series of downsampling and feature extraction layers, including average pooling, max pooling, and two residual blocks, capturing multi-scale luminance information.

Next, a \( 1 \times 1 \) convolutional layer generates filter weights that are adaptively adjusted based on the luminance information at each pixel. These pixel-wise filters, denoted \( h^x_i \), are then convolved with local image patches \( R(X_i) \) to produce the adaptive output features:

\begin{equation}
\hat{X}_i(x) = h^x_i \otimes R(X_i),
\end{equation}
where \( h^x_i \) is a filter with size \( u_p \times u_p \), ensuring each pixel position generates an adaptive filtering effect based on the surrounding luminance. This adaptive design enhances the model’s ability to handle complex lighting variations, facilitating more robust image information extraction.

The overall and pixel-by-pixel illumination adjustments should be jointly considered during training, as both global illumination enhancement and preservation of the original semantics during the brightness enhancement process are necessary. We use the loss functions that are well designed by Zero-IG~\cite{shi2024zero} expressed as:
\begin{equation}
L_{\text{LAM}} = L_{\text{over}} + L_{\text{pix}},
\end{equation}
$L_{\text{over}}$ represents the overall adjustment loss, $L_{\text{pix}}$ denotes the pixel-by-pixel adjustment loss. In this study, we adopt the same settings for the loss functions as those used in Zero-IG \cite{shi2024zero}. See the Supplemental Materials for more details.

\subsection{Reflect Augmentation Module}
In low-light action recognition, effectively capturing and utilizing illumination is crucial. Previous approaches typically handle illumination enhancement and action recognition separately, leading to suboptimal usage of available light. To address this, we introduce the Reflect Augmentation Module (RAM), featuring a dual-channel interactive illumination compensator designed to fully exploit ambient illumination directly within the recognition framework.

In RAM, each input feature $X_i$ undergoes deep processing through a dual-pathway structure. First, \( X_i \) is passed through the action recognition backbone of DSAR, which serves as the main pathway and is denoted by \( f_1(X_i) \). The output features from the main pathway are then passed to the reflected pathway, which has the same structure as the main pathway, but each stage retains only a single transformer block for lightweight processing. In the reflected pathway, the ``reflected features" capture latent global illumination dynamics, which are crucial for feature representation in low-light scenarios. The features from both the reflected and main pathways are then fused via combination \( \textcircled{c} \). This process not only fuses the features but also preserves spatially unique information from both two pathways. The entire process is defined as follows:
\begin{equation}
X_{i+1} = f_2(f_1(X_i)) \textcircled{c} f_1(X_i),
\end{equation}
where \( f_1 \) represents the core computation for the main pathway, \( f_2 \) denotes the reflected pathway, and \( \textcircled{c} \) represents the concatenation operation.

This architecture enables RAM to adaptively accumulate illumination-sensitive features across stages, facilitating robust action recognition under challenging low-light conditions. Each stage output \( X_{i+1} \) incorporates multi-scale illumination corrections, significantly enhancing the model's performance in dark environments.

The loss function for RAM is the standard cross-entropy loss, where \(y\) denotes the action category and \(\hat{y}\) is the predicted category, which is defined as: 
\begin{equation}
L_{RAM} = -\sum_{i} y_i \log(\hat{y}_i).
\end{equation}

Since our method is trained end-to-end, the total loss function is defined as:
\(L_{\text{total}} = L_{\text{TC}} + L_{\text{LAM}} + L_{\text{RAM}} \).

\subsection{Dark-101}
As we have described in the Introduction section that the existing low-light action recognition datasets have some limitations. ARID~\cite{xu2021arid} contains only 3,784 videos and ARID1.5~\cite{xu2021arid} expands to 5,572 videos, but both remain relatively small, with limited action categories and scenes. Recently, Dark-48~\cite{tu2023dtcm} enlarged the scale to 8,815 videos across 48 categories but included amounts of noise videos unrelated to actions, resulting in suboptimal training outcomes. 

To address these challenges and meet the demand of training deep learning big models, constructing a large-scale dataset is extremely urgent. However, building a completely new dark video dataset with a much larger number of videos and action categories is difficult. We collect useful dark videos from existing benchmarks economically and the Internet painstakingly. 

\begin{table}[htbp]
    \centering
    \begin{tabular}{lccc}
        \toprule
        \textbf{Dataset} & \textbf{Actions} & \textbf{Scenes} & \textbf{Total} \\ 
        \midrule
        ARID~\cite{xu2021arid} & 11 & 18 & 3784 \\ 
        ARID1.5~\cite{xu2021arid} & 11 & 24 & 5572 \\ 
        Dark-48~\cite{tu2023dtcm} & 48 & 52 & 8815 \\ 
        \midrule
        \textbf{Dark-101} & \textbf{101} & \textbf{138} & \textbf{18310} \\
        \bottomrule
    \end{tabular}
    \caption{Comparison of Dark Video Action Recognition Datasets: ARID~\cite{xu2021arid}, ARID1.5~\cite{xu2021arid}, Dark-48~\cite{tu2023dtcm}, and our Dark-101. `Actions' indicates the number of action classes, with each class comprising over 100 dark videos; `Scenes' represents the number of distinct scenes from which the videos are sourced; `Total' denotes the overall count of video clips.}
    \label{tab:duibi}
\end{table}

Specifically, to get valuable data from the existing dataset, we first defined an evaluation metric called Global Darkness Quantification (GDQ). It combines the mean brightness of video frames with intra-frame brightness fluctuation to produce a quantified darkness index, distinguishing low-light videos from the normal-light ones.

For each video frame $I_t(x, y)$, $(x, y)$ are pixel coordinates, the mean brightness $\mu_t$ is calculated as:
\begin{equation}
\vspace{-0.1cm} 
\mu_t = \frac{1}{M \cdot N} \sum_{x=1}^{M} \sum_{y=1}^{N} I_t(x, y),
\end{equation}
where $M$ and $N$ represent the frame’s height and width.

To obtain a global brightness baseline, we calculate the average brightness $\mu_c$ across all video frames $T$:
\vspace{-0.1cm} 
\begin{equation}
\mu_c = \frac{1}{T} \sum_{t=1}^{T} \mu_t.
\end{equation}
The darkness index $D_v$ quantifies a video’s darkness by combining the difference between each frame’s brightness $\mu_t$ and the global average $\mu_c$ with the intra-frame brightness fluctuation $\sigma_t$ (the standard deviation), which is defined as:
\vspace{-0.1cm} 
\begin{equation}
D_v = \frac{1}{T} \sum_{t=1}^{T} \left( \frac{\mu_t - \mu_c}{\mu_c} \right) \cdot \sigma_t.
\label{eq:pingjia}
\end{equation}
The video is classified as low-light if $D_v < -\tau$, otherwise as normal-light. Where $\tau$ is determined by the average darkness across all videos in ARID1.5 as~\cite{tu2023dtcm} ($\tau = 0.877$).

To align with the darkness of the ARID1.5 dataset, we first evaluate the darkness of each video via our designed video darkness evaluation measure Eq.~\ref{eq:pingjia}. Moreover, we crawl a substantial amount of video data from the Internet publicly. The sources encompass platforms like YouTube. Specifically, among all the data we collected, 2689 videos are captured from the Internet and 15621 videos are selected from existing datasets, including Kinetics700 \cite{carreira2019short}, Dark-48 \cite{tu2023dtcm}, Something-Something V3 \cite{goyal2020something} and MIT \cite{monfort2019moments}. We count the categories of the selected dark videos and retain only those categories with more than 150 videos, helping to solve the problem of long-tail distributions.

Compared with the previous datasets, our dataset exhibits a significantly lower illumination levels, a substantially larger number of videos, a more extensive variety of action categories, and a greater diversity of video scenarios. Besides, the duration of the videos ranges from 5 to 10 seconds, which is longer than that of the previous datasets (approximately 3 seconds). Consequently, it is possible to train models that with more powerful temporal extraction capability, and larger-size models that are not only more robust but also more generalized in application. The training and testing sets are partitioned by assigning 80\% of the videos to the training set and the remaining 20\% to the testing set. A statistical comparison among Dark-101, ARID, ARID1.5~\cite{xu2021arid}, and Dark-48~\cite{tu2023dtcm} is given in Table~\ref{tab:duibi}. More details can be seen in the Supplemental Materials.
\section{Experiment Results}
\subsection{Implementation Details}

\textbf{Parameter Settings.} All experiments are implemented with PyTorch on 8 Nvidia 4090 GPUs. The ADAM optimizer is employed, with parameters $\beta_1 = 0.9$, $\beta_2 = 0.999$, and $\epsilon = 10^{-8}$. The batch size is set to 8 and learning rate is fixed at $2 \times 10^{-4}$. The training epoch number is set to 100.

\begin{table*}[ht]
\centering
\begin{tabular}{lccc}
\hline
\textbf{Model} & \textbf{Pretrained} & \textbf{Top-1 (\%)} & \textbf{Top-5 (\%)} \\
\hline
TSN~\cite{wang2016temporal} & K400 & 34.04 & 88.29  \\
TSM~\cite{lin2019tsm} & K400 & 61.37 & 91.80  \\
SlowFast~\cite{feichtenhofer2019slowfast} & K400 & 66.76 & 91.58  \\
R(2+1)D~\cite{tran2018closer} & K400 & 63.52 & 92.29  \\
Swin-T~\cite{liu2022video} & K400 & 67.69 & 97.55  \\
TimeSformer~\cite{bertasius2021space} & K400 & 49.36 & 93.17  \\
UniFormerV2~\cite{li2022uniformerv2} & K400 & 61.41 & 96.04  \\
MViTv2-B~\cite{li2022mvitv2} & K400 & 87.68 & 98.55   \\
Darklight+R(2+1)D-34~\cite{chen2021darklight} & IG65M & 84.13 & 97.34  \\
DTCM$^*$~\cite{tu2023dtcm} & K400 & 86.42 & 97.28 \\
R(2+1)D-GCN+BERT~\cite{singh2022action} & IG65M & 86.93 & \textbf{\textcolor{blue}{99.35}}\\
Dark-DSAR~\cite{yin2024dark} & K400 & \textbf{\textcolor{blue}{89.49}} & 98.77  \\
\textbf{OwlSight} & \textbf{K400} & \textbf{\textcolor{red}{94.85}} & \textbf{\textcolor{red}{99.50}}\\
\hline
\end{tabular}
\caption{Comparison with the state-of-the-arts on the ARID1.5~\cite{xu2021arid} dataset. All results are calculated using the RGB channel. \textcolor{red}{Red} and \textcolor{blue}{blue} colors indicate the best and second-best performance, respectively. The superscript $^*$ indicates the model is retrained on ARID1.5.}
\label{tab:table1}
\end{table*}

\textbf{Compared Methods.} We compare our method OwlSight with several state-of-the-art low-light action recognition methods, e.g. Dark-DSAR~\cite{yin2024dark}, DTCM~\cite{tu2023dtcm}, DL-KDD~\cite{chang2024dl}, R(2+1)D-GCN+BERT~\cite{singh2022action}, Darklight+R(2+1)D-34~\cite{chen2021darklight}. The results are reproduced using the publicly available source codes with the recommended parameters.

\textbf{Benchmarks Description and Metrics.} We conduct experiments on four representative dark video action recognition datasets: ARID~\cite{xu2021arid}, ARID1.5~\cite{xu2021arid}, Dark-48~\cite{tu2023dtcm}, and our Dark-101. We evaluate the performance via Top-1 and Top-5 accuracy. For ARID1.5, both Top-1 and Top-5 are reported due to its being the most widely used dark video dataset. For other datasets, we report only Top-1 accuracy. Dark action recognition results for relevant methods are taken from their original papers, while for those without available results, we reproduce them (marked with *).
\vspace{-0.1cm}
\subsection{Benchmark Evaluations}
\textbf{Performance Evaluation.} Our method OwlSight is evaluated with 32-frame inputs, sampled at 3-frame intervals, using standard data augmentation techniques like random resized cropping, horizontal flipping, and random erasing. In quantitative comparisons, our method outperforms others on ARID1.5 and ARID datasets, as presented in Table~\ref{tab:table1} and \ref{tab:table2}. Furthermore, on both Dark-48 and Dark-101 datasets, our method demonstrates superior performance over other existing algorithms. Due to space limitation, more comparisons are available in the Supplemental Materials.
\begin{table}[htbp]
    \centering
    \begin{tabular}{lc}
        \hline
        \textbf{Model} & \textbf{Top-1 (\%)} \\
        \hline
        Pseudo-3D-199~\cite{qiu2017learning} & 71.93 \\
        UniFormerV2~\cite{li2022uniformerv2} & 73.39 \\
        TimeSformer-L~\cite{bertasius2021space} & 81.39 \\
        Video-Swin-B~\cite{liu2022video} & 89.79 \\
        MViTv2-B~\cite{li2022mvitv2} & 91.43 \\
        DarkLight-R(2+1)D-34~\cite{chen2021darklight} & 94.34 \\
        Suman~\cite{suman2023two} et al. & 95.86 \\
        Dark-DSAR~\cite{yin2024dark} & 96.99 \\
        DL-KDD~\cite{chang2024dl} & 97.27 \\
        \rowcolor[rgb]{0.9,0.9,0.9} 
        \textbf{OwlSight} & \textbf{99.27}\\
        \hline
    \end{tabular}
    \caption{Comparison with state-of-the-arts on ARID.}
    \label{tab:table2}
\end{table}
\vspace{-0.5cm}

\subsection{Ablation Study}

To evaluate the impact and performance of each component in our \textit{OwlSight}, we examine their function in this section. For testing, we select the representative dataset ARID1.5.

\textbf{Effect of Each Component.} As shown in Table \ref{table:6}, the \textit{Baseline} model is obtained by removing the LAM, RAM (only the reflected pathway) and TCM from our OwlSight, without applying any additional optimizations. The function of the LAM is to adaptively adjust the input low-light video to normal illumination based on the input brightness range. When the LAM is removed, accuracy decreases by 2.45\%, from 94.85\% to 92.4\%. Which demonstrates the adaptive illumination enhancement of LAM is effective. 
When the RAM is removed, accuracy drops by 3.03\%, from 94.85\% to 91.82\%. This is because the designed RAM aims to interactively exploit illumination information for better recognition. Without it, OwlSight loses one pathway for leveraging the illumination information. 
When the TCM is removed, the results show a 1.81\% reduction in accuracy, from 94.85\% to 93.04\%. Which reveals preserving temporal coherence across the enhanced frame sequence is helpful for dark video recognition. 


\begin{table}[ht]
\setlength{\tabcolsep}{5pt}
\centering
\begin{tabular}{lcccc}
\hline

&\textbf{LAM} & \textbf{RAM} & \textbf{TCM} & \textbf{Top-1 (\%)} \\
 \hline
 
Baseline&\ding{55}& \ding{55}& \ding{55}&87.68\\ 
  &\ding{55}    & \ding{51}    & \ding{51}    & 92.40\\  
  &\ding{51}    & \ding{55}    & \ding{51}    & 91.82\\ 
  &\ding{51}    & \ding{51}    & \ding{55}    & 93.04\\ 
\rowcolor[gray]{0.9} 
  OwlSight&\ding{51}    & \ding{51}    & \ding{51}    & 94.85\\ \hline
\end{tabular}
\caption{Ablation study of each component in our model OwlSight. The Top-1 Accuracy on ARID1.5 is reported.}
\label{table:6}
\vspace{0.2cm}
\centering
\begin{tabular}{ccccc}
\hline
\textbf{$L_{ar}$} & \textbf{$L_{TC}$} & \textbf{$L_{over}$} & \textbf{$L_{pix}$} & \textbf{Top-1 (\%)} \\ \hline
\rule{0pt}{10pt} 
\ding{51}         & \ding{55}         & \ding{51}          & \ding{51}          & 92.03               \\ 
\ding{51}         & \ding{51}         & \ding{55}          & \ding{51}          & 91.32              \\ 
\ding{51}         & \ding{51}         & \ding{51}          & \ding{55}          & 91.25               \\ 
\rowcolor[gray]{0.9} 
\ding{51}         & \ding{51}         & \ding{51}          & \ding{51}          & 94.85               \\ \hline
\end{tabular}
\caption{Ablation study of the loss terms in our OwlSight.}
\label{table:7}
\end{table}

\begin{figure*}[ht]
    \centering
    \hspace{-0.6cm} 
    \includegraphics[width=1\textwidth]{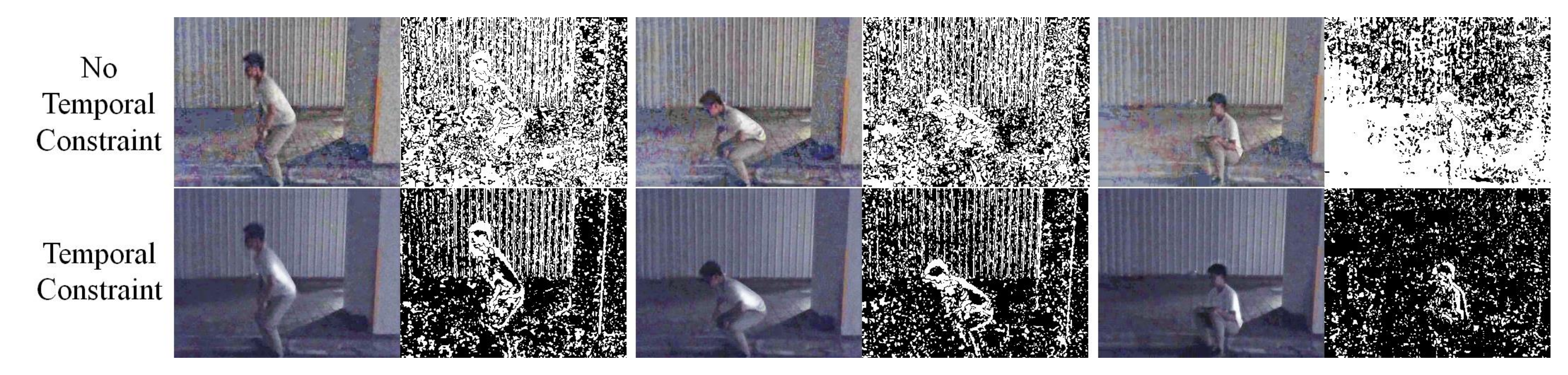}
    \caption{%
        Comparison of RGB-Difference of our OwlSight with and without temporal constraint (TCM and $L_{TC}$) on the ARID1.5 dataset. Results are as follows:
        The upper part shows the RGB-Difference of OwlSight without temporal constraint, which are failed to improve the action recognition performance in dark videos. The lower part shows the RGB-Difference of OwlSight with temporal constrain. Obviously, the introduced temporal constraint strategy significantly boosts the performance of action recognition in dark videos.
    }
    \label{fig:time_constrain}
\end{figure*}

\vspace{-0.1cm}
\textbf{Effect of Loss Function.} We explore the contribution of our loss function in detail. Since the cross-entropy \( L_{ar} \) used by \( L_{RAM} \) determines the final classification result, it cannot be removed. As shown in Table~\ref{table:7}, the Top-1 accuracy decreases from 94.85\% to 91.25\% after removing the pixel-by-pixel adjustment loss (\( L_{pix} \)). Furthermore, when the overall adjustment loss (\( L_{over} \)) is removed, the accuracy decreases by 3.53\%, from 94.85\% to 91.32\%. The temporal consistency loss (\( L_{tc} \)) is employed to constrain the Time-Consistency module better, thereby extracting more effective features in the temporal dimension, and when it is removed, the accuracy decreases by 2.82\%, dropping from 94.85\% to 92.03\%. Finally, when combining all the loss functions with our method, the model achieves optimal performance on accuracy metrics, which demonstrates that our loss functions are reasonable. More ablation experiments are included in the Supplemental Materials.

\textbf{Parameter Setting Analysis.}
As shown in Table~\ref{table:8}, to determine the optimal parameter setting of \( \mu_{\text{out}} \), we examine how the average brightness of the normal-light image on ARID1.5, highlighting that \( \mu_{\text{out}} = 0.5 \) achieves the highest Top-1 accuracy.

\begin{table}[ht]
\vspace{-0.2cm}
\centering
\begin{tabular}{cccccc}
\hline
 \( \mu_{\text{out}} \) & 0.3 & 0.4 & 0.5 & 0.6 & 0.7 \\ \hline
Top-1 (\%) & 90.32 & 92.23& \textbf{94.85} & 94.23 & 92.76 \\ \hline
\end{tabular}
\caption{Accuracy of Different \( \mu_{\text{out}} \) Configurations on ARID1.5.}
\label{table:8}
\vspace{0.2cm}

    \centering
    \begin{tabular}{ccc}
        \hline
        & \textbf{$h^x_i \ast X$} & \textbf{$h^x_i \ast Y$} \\ 
        \hline 
        Top-1 (\%) & 94.85 & 89.64 \\
        \hline
    \end{tabular}
    \caption{Comparison of recognition performance using generated kernels with input $X$ vs. $Y$. }
    \label{table:10}
    \vspace{-0.2cm}
\end{table}

\textbf{LAM Analysis.}
The LAM module uses the generated convolution kernels $h^x_i$ to convolve with the input $X$ to obtain the enhanced image instead of directly using $Y$. This is because the Gamma transformation has inherent limitations. Its global monotonic mapping makes it difficult to handle multi-scale light distortion and preserve high-frequency details.
Our experimental results in Table~\ref{table:10} show that fusing the generated kernels with the input's spatiotemporal features significantly outperforms directly convolving with $Y$ in recognition performance.

\subsection{Qualitative Results}

\textbf{Visual Analysis of Temporal Constraint Effectiveness.} We compare the RGB-Differences of our OwlSight with and without temporal constraint (TCM and $L_{TC}$). As Figure~\ref{fig:time_constrain} shows, method in the upper part, which generates confused RGB-Differences (chaotic and unclear), fails to improve the action recognition performance in dark videos. Method in the bottom part indicates that our OwlSight, which generates clear RGB-Differences, succeeds in enhancing the action recognition performance in dark videos. The clean static background and clear motion boundaries demonstrates that the spatio-temporal consistency is well preserved by the introduced temporal constraint strategy. In summary, only the dark enhancement method, which can maintain the temporal consistency of dark videos, is beneficial to dark video action recognition. More analysis experiments are included in the Supplemental Materials.

\vspace{-0.3cm} 
\section{Conclusion}
\label{sec:conclusion} 
\vspace{-0.2cm}
We propose a novel biomimetic design-based framework OwlSight, for low-light action recognition by fully exploiting the brightness information. OwlSight comprises three key components: lightweight TCM, LAM, and RAM. The TCM captures the spatiotemporal relationships within dark videos. LAM dynamically adjusts the global brightness based on the input feature illumination. RAM enhances action recognition by maximizing the available illumination information. In addition, we construct a large-scale new dataset Dark-101, comprising 18,310 low-light video samples with 101 action categories. Extensive experiments demonstrate that our OwlSight achieves the outstanding performance for diverse datasets with a significant gain in accuracy compared to the prior methods.

{
    \small
    \bibliographystyle{ieeenat_fullname}
    \bibliography{main}
}

\clearpage
\setcounter{page}{1}
\appendix
\maketitlesupplementary

\section{\texorpdfstring{Description of the $L_{TC}$ and the $L_{LAM}$}{Description of the L\_TC and the L\_LAM}}
\subsection{\texorpdfstring{Details of $L_{TC}$}{Details of L\_TC}}
During training phase, we employ a temporal consistency loss~\cite{tu2023dtcm}, denoted as \( L_{\text{TC}} \). It encourages the enhanced features to closely mimic the temporal variations present in the original dark video clips. Here, \( K \) represents the number of local regions, and \( i \) refers to the eight neighboring regions (top, bottom, left, right, top-left, top-right, bottom-left, bottom-right) surrounding the region \( i \). The variable \( T \) signifies the total number of frames in the input video clip, while \( D Y \) and \( D I \) correspond to the enhanced frame and the input frame, respectively. Notably, \( \left| D Y_{t+1} - D Y_t \right| \) represents the RGB difference between consecutive enhanced video frames, and \( \left| D I_{t+1} - D I_t \right| \) denotes the RGB difference between consecutive dark video frames.

The spatial consistency loss \( L_{\text{SCF}}(DY, DI) \) is calculated as:
\begin{equation}
L_{\text{scf}}(D Y, D I) = \frac{1}{K} \sum_{i=1}^{K} \sum_{j \in \Omega(i)} \left( \left| Y_i - Y_j \right| - \left| P_i - P_j \right| \right)^2,
\end{equation}
where \( \Omega(i) \) denotes the set of indices for the eight neighboring regions of region \( i \).

The temporal consistency loss \( L_{\text{TC}} \) is then formulated as:
\begin{equation}
L_{\text{TC}} = \frac{1}{T - 1} \sum_{t=1}^{T - 1} L_{\text{scf}} \left( \left| D Y_{t+1} - D Y_t \right|, \left| D I_{t+1} - D I_t \right| \right),
\end{equation}
where \( L_{\text{tcf}} \) measures the difference between the temporal changes of the enhanced and the input video frames.

\subsection{\texorpdfstring{Details of $L_{LAM}$}{Details of LAM}}
The loss function for our LAM is defined as: 
\begin{equation}
L_{\text{LAM}} = L_{\text{over}} + L_{\text{pix}}.
\end{equation}
The overall adjustment loss, represented by \( L_{\text{over}} \), constrains illumination by adjusting image brightness, enabling proportional upscaling of all pixels in \( I \). The adjusted image can be expressed as \( I \circ S^{-1} = \alpha I \). Here, the symbol \(\circ\) denotes element-wise multiplication. \(\alpha\) is the brightness coefficient, defined as \( \alpha = Y_H Y_L^{-1} \), where \( Y_H \) represents the mean value of the luminance plane \( Y \) of normal-light images, using a reference value of 0.5 as Zero-IG~\cite{shi2024zero}. \( Y_L \) indicates the mean value of the luminance plane \( Y \) of \( I \).
\begin{equation}
L_{\text{over}} = \| S - \alpha^{-1} \|_2^2
\label{eq:4}
\end{equation}
The pixel-by-pixel adjustment loss, \( L_{\text{pix}} \), constrains the illumination based on the intensity of each pixel in \( I \). The scaling factor \(\beta\) controls the scaling degree, thereby adjusting the level of contrast enhancement. Consistent with Zero-IG~\cite{shi2024zero}, we set \(\beta = 0.7\), and \( L_{\text{pix}} \) is defined as:

\begin{equation}
L_{\text{pix}} = \| S - \beta (\alpha I)^\alpha \|_2^2
\label{eq:5}
\end{equation}

\section{Dataset Description}

We collect the useful dark videos from current benchmark datasets and Internet. To align with the darkness of the ARID1.5 dataset:
\begin{enumerate}
    \item We evaluate the darkness of every ARID1.5 video by using our designed video darkness evaluation measure Eq.~\ref{eq:pingjia}, and set $\tau$ to be the average darkness of all ARID1.5 videos.
    \item We access the darkness of the videos from datasets such as Kinetics700 \cite{carreira2019short}, Dark-48 \cite{tu2023dtcm}, Something-Something V3 \cite{goyal2020something}, and MIT \cite{monfort2019moments}, as well as some videos sourced from the Internet (e.g., YouTube). We then select the dark videos according to Eq.~\ref{eq:pingjia}.
\end{enumerate}
\begin{figure}[ht]
    \centering
    \hspace{-0.6cm} 
    \includegraphics[width=0.4\textwidth]{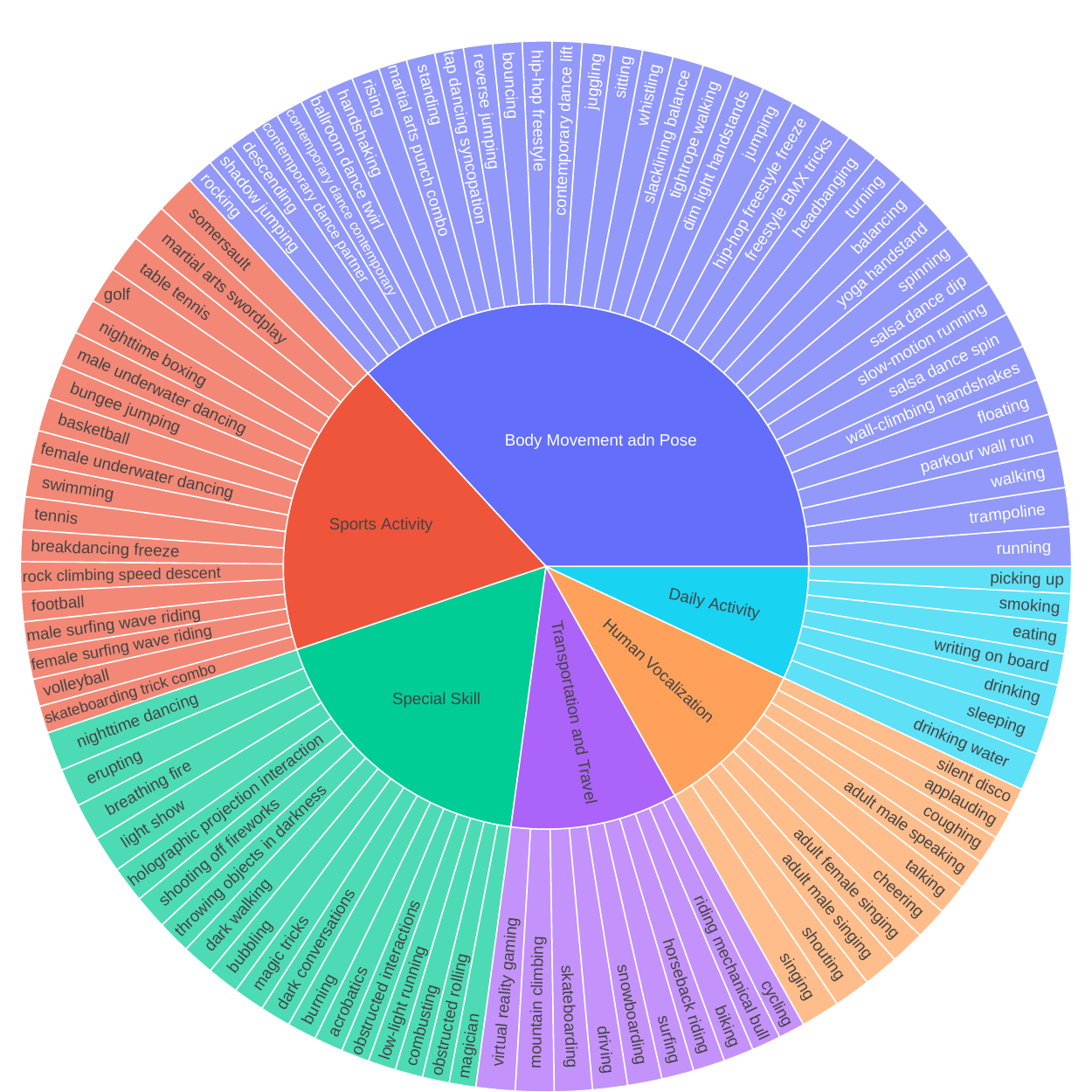}
    \caption{The video distribution for all action classes in the constructed dataset Dark-101. The ratio of the training set to the testing set is 4:1. Each action class in our Dark-101 contains at least 150 dark videos.}
    \label{fig:circle}
\end{figure}
\begin{figure}[ht]
    \centering
    \hspace{-0.6cm} 
    \includegraphics[width=0.4\textwidth]{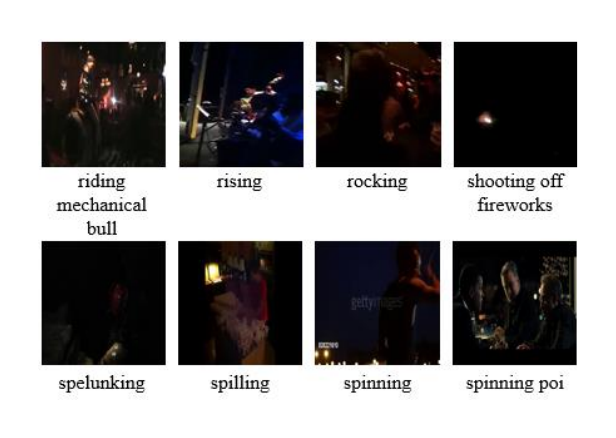}
    \caption{Example action classes from our Dark-101 datasets.}
    \label{fig:data}
\end{figure}

Our dark video dataset Dark-101 contains 18310 dark videos belonging to 101 action categories. Some example classes in extremely dark conditions from our Dark-101 dataset are shown in Figure~\ref{fig:data}. And the visualization of the distribution of the number of datasets can be further viewed in the externally provided HTML file by clicking action items. 

\section{Quantitative Comparisons}
\subsection{Dark48}
\textbf{Evaluation on the Dark-48~\cite{tu2023dtcm} dataset.} Dark-48~\cite{tu2023dtcm} provides a larger and more diverse collections than ARID and ARID1.5, featuring 48 action categories and 8,815 video clips, which enhances the range of testing conditions in low-light scenarios. Our proposed OwlSight achieves the highest accuracy $48.24\%$, surpasses the recent approaches by a margin of at least $1.56\%$.
\begin{table}[ht]
\centering

\begin{tabular}{lc}
\hline
\textbf{Model} & \textbf{Top-1 (\%)} \\
\hline
TSN~\cite{wang2016temporal} & 26.04 \\
TSM~\cite{lin2019tsm} & 36.46 \\
TIN ~\cite{shao2020temporal}& 33.38 \\
3D-ResNext-101~\cite{hara2018can} & 37.23 \\
SlowFast16+64 ~\cite{feichtenhofer2019slowfast}& 39.58 \\
TimeSformer-L~\cite{bertasius2021space} & 43.27 \\
Video-Swin-B~\cite{liu2022video} & 41.92 \\
MViT-B~\cite{li2022mvitv2} & 40.37 \\
Dark-DSAR~\cite{yin2024dark} & 42.33  \\
DarkLight-ResNext101~\cite{chen2021darklight} & 42.27 \\
DTCM$^*$~\cite{tu2023dtcm} & 46.68 \\
\rowcolor[gray]{0.9} 
\textbf{OwlSight} & \textbf{48.24} \\
\hline
\end{tabular}
\caption{Comparison with state-of-the-arts on Dark-48~\cite{tu2023dtcm}.}
\label{tab:table3}
\end{table}

\subsection{Dark101}
\textbf{Evaluation on our Dark-101 Dataset.} These existing datasets are small-scale with insufficient number of categories, causing they are difficult to capture the diversity and complexity needed for training large-scale low-light action recognition models. To promote real-world low-light action recognition research and test the generalization capacity of the proposed OwlSight, a comprehensive comparison is conducted on our new dataset \textit{i.e.} Dark-101. As shown in Table~\ref{tab:table4}, our OwlSight obtains an improvement of 1.72\% over the other models, due to that it can fully exploit the illumination information and easily adapts to larger and more diverse dark action recognition datasets, meaning that it possesses stronger generalization capacity.

\begin{table}[ht]
\centering
\begin{tabular}{lc}
\hline
\textbf{Model} & \textbf{Top-1 (\%)} \\
\hline
TimeSformer-L$^*$~\cite{bertasius2021space} & 36.39 \\
Swin-B$^*$~\cite{liu2021swin} & 37.79 \\
Video-Swin-B$^*$~\cite{liu2022video} & 40.79 \\
UniFormerV2$^*$~\cite{li2022uniformerv2} & 42.39 \\
MViTv2-B$^*$~\cite{li2022mvitv2} & 47.43 \\
DTCM$^*$~\cite{tu2023dtcm} & 50.91 \\
Dark-DSAR$^*$~\cite{yin2024dark} & 52.13 \\
\rowcolor[gray]{0.9} 
\textbf{OwlSight} & \textbf{53.85} \\
\hline
\end{tabular}
\caption{Comparison with state-of-the-arts on our Dark-101.}
\label{tab:table4}
\end{table}

\section{RAM Analysis.}

Since RAM performs exceptionally well, to demonstrate that the enhancement of the experimental results stems from the principle of the module we designed rather than the enlargement of the number of model parameters, we carried out an ablation experiment regarding the number of parameters of RAM. Specifically, when the RAM module is removed, the total number of parameters in the entire model is 51.4M. When the RAM module is incorporated, the total number of parameters in the entire model is 54.7M, where an increasing of merely 3.3M. However, the Top-1 accuracy increased by 3.03\%. This experimental outcome verifies that the improvement in our experimental results is not attributed to the expansion of the parameter scale, as shown in Table~\ref{table:9}.

\begin{table}[ht]
\centering
\begin{tabular}{ccc}
\hline
 & \textbf{w/} & \textbf{w/o} \\ 
\hline 
Top-1 (\%) & 94.85 & 91.82 \\
Params (M) & 54.7 & 51.4 \\
\hline
\end{tabular}
\caption{Ablation study of RAM on the ARID1.5 dataset. The results show the impact of including and excluding the RAM module on the model's performance and parameter count.}
\label{table:9}
\end{table}





\begin{figure}[ht]
    \centering
    \hspace{-0.6cm} 
    \includegraphics[width=0.5\textwidth]{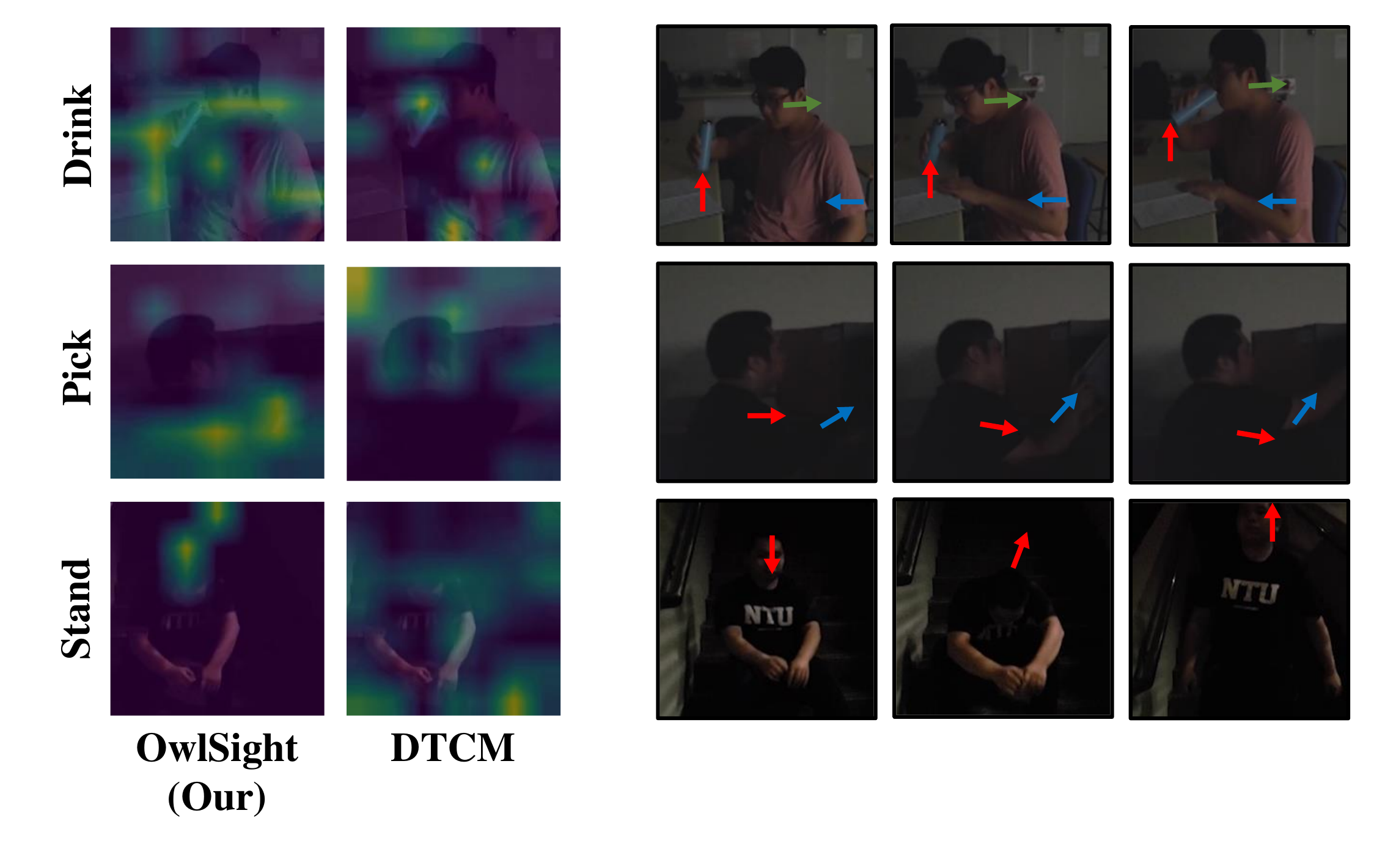}
    \vspace{-0.5cm}
    \caption{The left figure displays the Grad-CAM visualizations for three selected actions (``Drink", ``Pick" and ``Stand") from the ARID1.5 dataset. The gradients for the Grad-CAM maps are extracted from the second-to-last layer of our OwlSight and DTCM. The right side illustrates the full sequence of actions, with arrows indicating the movement trends of different body parts.}
    \label{fig:grad-cam}
\end{figure}

\section{Visualize the attention.}
As shown in Figure~\ref{fig:grad-cam}, we visualize the attention of our OwlSight and compare it with DTCM~\cite{tu2023dtcm}. The selected actions, ``Drink," ``Pick," and `Stand" from the ARID1.5 dataset, are used for comparison. In the Grad-CAM map (left side of Figure~\ref{fig:grad-cam}), regions with action movements are highlighted. For the ``Drink" action, it is clear that our OwlSight focuses on motion changes with higher illumination compared to DTCM. For the ``Pick" action, our OwlSight provides a more accurate depiction of the motion changes, whereas DTCM tends to focus more on the general scene. For the ``Stand" action, our OwlSight effectively captures the primary movement, while DTCM includes a significant amount of irrelevant background information. 

On the right side of Figure~\ref{fig:grad-cam}, the entire action sequence is displayed, with arrows indicating the movement trends of different body parts. For example, the blue arrow in the ``Pick" action highlights the arm movement, which is given significant attention. These results demonstrate that our model can accurately capture motion changes in low-light scenarios meanwhile reducing focus on irrelevant areas.




\end{document}